%% file: 00_main.tex
\documentclass[a4paper]{article}
\usepackage{INTERSPEECH2021}
\usepackage{subcaption}
\usepackage{multirow, makecell}
\usepackage{xcolor}
\usepackage{pifont}
\usepackage{amssymb}
\usepackage{amsmath}
\usepackage{cite}
\usepackage{hyperref}
\usepackage{enumitem}
\usepackage{tikz}
\usetikzlibrary{patterns}
\usepackage{pgfplots}
\pgfplotsset{compat=1.7}

\usepackage[colorinlistoftodos]{todonotes}
\usepackage{booktabs}
\usepackage{fdsymbol}
\newcommand{\paragraphHdTop}[1] {\noindent\textbf{#1}}

\newcommand\aspace{\hspace{.75em}}

\title{Rethinking End-to-End Evaluation of Decomposable Tasks:
A Case Study on Spoken Language Understanding }
\name{Siddhant Arora${}^*$ \aspace\aspace Alissa Ostapenko${}^*$ \aspace\aspace Vijay Viswanathan${}^*$ \aspace\aspace Siddharth Dalmia${}^*$\\
Florian Metze \aspace\aspace Shinji Watanabe \aspace\aspace Alan W Black}
\address{Language Technologies Institute, Carnegie Mellon University, USA}
\email{\{siddhana,aostapen,vijayv,sdalmia\}@cs.cmu.edu}

\begin{document}
\ninept

\maketitle
\begin{abstract}
%Human-interfacing tasks are complex, involving several hierarchical sub-tasks that each test a different model capability.
Decomposable tasks are complex and comprise of a hierarchy of sub-tasks.  Spoken intent prediction, for example, combines automatic speech recognition and natural language understanding. Existing benchmarks, however, typically hold out examples for  only  the  surface-level  sub-task.   As  a  result,  models with similar performance on these benchmarks may have unobserved  performance  differences  on  the  other  sub-tasks.   To allow insightful comparisons between competitive end-to-end architectures, we propose a framework to construct robust test sets using coordinate ascent over sub-task specific utility functions. Given  a  dataset  for  a  decomposable  task,  our  method optimally creates a test set for each sub-task to  individually  assess  sub-components of the end-to-end model. Using spoken language understanding as a case study, we generate new splits for the Fluent Speech Commands and Snips SmartLights datasets. Each split has two test sets:  one with held-out utterances assessing natural language understanding abilities, and one with held-out speakers to test speech processing skills.  Our splits identify performance gaps up to 10\% between end-to-end systems that were within 1\% of each other on the original test sets.  These performance gaps allow more realistic and actionable comparisons between different architectures, driving future model development. We release our splits and tools for the community.\footnotemark \let\thefootnote\relax\footnotetext{${}^*$Equal Contribution.${}^1$We call our software package MASE (Multi-Aspect Subtask Evaluation): \url{https://MASEeval.github.io/}.}

%   Our work analyzes and extends a recently published state-of-the-art system for Spoken Language Understanding (SLU) of commands typically used with voice assistants. Through careful data analysis and experimentation, we show that their model, achieving 98.5\% accuracy in mapping a spoken command to its intents, is unable to generalise to new commands. By extending their model with semantic word embeddings in addition to speech features, we report 5\% gains in performance when generalising to new utterances.
\end{abstract}
\input{01_introduction}

\input{01_task}
\input{02_methods}
\input{03_results}
\input{04_conclusion}
\bibliographystyle{IEEEtran}
\footnotesize{ \bibliography{mybib}}
\end{document}

%% file: 01_introduction.tex
\section{Introduction}
Complex, real-world tasks, such as the spoken language understanding (SLU) tasks of spoken intent prediction and spoken language translation, comprise of hierarchies of simpler sub-tasks. Spoken intent prediction combines automatic speech recognition (ASR) to process audio, followed by natural language understanding (NLU) to classify an utterance to a particular intent (intent prediction) \cite{gorin1997may}.  Similarly, speech translation involves an ASR task followed by machine translation (MT) to translate a transcription of the input audio  \cite{weiss2017sequence}.

Deep, end-to-end models \cite{qian2021speechlanguage,lugosch2019speech,Huang_E2E,Kuo2020,Chen_E2E,Dinarelli_E2E} are adopted for these complicated tasks due to advancements in model architectures and computing capabilities. End-to-end architectures typically outperform traditional, modular architectures without requiring domain expertise or feature engineering \cite{Agrawal2020TieYE}. Moreover, end-to-end models avoid the error propagation arising from traditional approaches \cite{sperber2020speech}. However, traditional modular or cascade architectures, naturally structured into sub-components that each address a specific sub-task, are more straightforward to evaluate. End-to-end models cannot quantify performance of decomposed sub-tasks \cite{dalmia-et-al}, blurring the lines between the individual sub-tasks.  Using pre-trained systems for some sub-tasks further reduces the chance of errors propagating to the downstream sub-tasks \cite{pretrain1, pretrain2, pretrain3}. Thus, it is important to explicitly evaluate each sub-component of an end-to-end network.

Prior datasets for decomposable problems, however, often test only the top-level subtask. For example, the Fluent Speech Commands (FSC) \cite{lugosch2019speech} and Air Travel Information System (ATIS) \cite{hemphill1990atis} SLU benchmarks are open-speaker but not open-utterance, and thus, only effectively test speaker generalizability. Moreover, train-test overlap is a problem in modern question answering datasets \cite{lewis2020question}. In paradigms like encoder-decoder modeling or speech translation \cite{dalmia-et-al,Speech_Translation,S2STranslate_google,S2S_Europe}, neural network components each solve different logical functions which combine to solve the final task. Therefore, standard benchmarks may effectively test a particular sub-network of a given system, masking any weaknesses of other model sub-components and providing an inflated estimates of model performance. 
%Thus, it is important to create benchmark test sets that evaluate different components of a learned model by assessing it across \textit{all} dimensions of variation in the data.

To address this, we present a dataset-agnostic framework for evaluating end-to-end model on decomposable tasks. Using spoken intent prediction as a case study, we focus on two popular benchmarks,  FSC \cite{lugosch2019speech} and Snips SmartLights (Snips) \cite{Saade2018SpokenLU,coucke2018snips} datasets. We provide evidence that the original test splits do not fairly evaluate the ASR and NLU subtasks of spoken intent prediction. Using our framework, we propose robust \textit{Unseen} and \textit{Challenge} splits that each contain two test sets: one test set with held-out speakers, and one with held-out utterances. For the Challenge set, we use coordinate ascent with speaker- and transcript-specific utility metrics to explicitly test for generalization to diverse speakers and varied phrasings of intents. Our experiments show the new test splits can amplify accuracy differences by up to 10\% between sub-components of several state-of-the art models for spoken intent prediction, offering more in-depth analyses of strengths and weaknesses of various end-to-end modeling approaches. These splits have the potential to drive future modeling innovation, not only for this task, but for any similarly decomposable task.

%% file: 01_task.tex
\section{Motivation}

In the following section, we introduce the spoken intent prediction task and discuss limitations of existing SLU benchmarks. 
\subsection{Task Definition}
\label{subsection:task}
Spoken intent prediction maps a spoken command (e.g. ``Turn on the lights in the kitchen") and to a discrete, actionable set of slots (Action: ``Activate"), (Object: ``Lights"), (Location: ``Kitchen"). This task challenges a model's speech recognition and semantic processing abilities: a good SLU model must generalize to new speakers \textit{and} to new phrasings of similar intents. 

The FSC dataset \cite{lugosch2019speech} tests a model's ability to predict intents from commands used with a home voice assistant. Following traditional speech processing paradigms, the FSC test set consists of audio from speakers unseen during training. Although this test split measures generalization to new speakers, the test set fails to explicitly test generalization to new utterances. As Table \ref{tab:model-comparison-fsc} illustrates, the training set provides 100\% coverage of the transcripts seen during test time. 
% This issue with transcript coverage is also seen in the popular ATIS  benchmark \cite{hemphill1990atis} for slot filling from air travel commands. 
Snips, which is similar in content to FSC, does not release official splits, but typical split creation approaches \cite{coucke2018snips,  qin2019stack, Agrawal2020TieYE} do not explicitly consider testing generalizability for each sub-task independently and instead use random splits which can lead to overtly optimistic estimate\cite{Talk_random_splits}. This can also mask performance gaps in thesub-tasks.

% \small{
\begin{table}[t]
% \footnotesize
\centering
\caption{\% WER values for Google ASR \cite{GoogleASR} on speakers with first language English and non English. (S, I, and D refer to substitution, insertion and deletion errors, respectively)}
\label{tab:native-wer-value}
\begin{tabular}{lcccc}
\toprule
First Language Spoken & \% S & \% I & \% D & \% WER \\ \midrule
English &  2.0 & 0.7 & 2.4 & 10.8 \\ 
Non English & 6.3 & 2.2 & 7.7 & 27.8 \\ \bottomrule
\end{tabular}
\vspace{-10px}
\end{table}
% }
In the real world, we expect systems to understand the same commands spoken in different ways by speakers of diverse backgrounds. Thus, it is important to hold out new utterances to more robustly assess a model's semantic processing ability \cite{semcomplex2020}. Moreover, open-speaker test sets should assess model generalizability to diverse demographics. In FSC, for example, all held-out speakers are native English speakers, while accented speakers are seen only during training time. To understand how this affects spoken language understanding evaluation, we used Google's ASR system \cite{GoogleASR} to generate transcripts from audio files in the dataset and computed Word Error Rates using the gold transcripts. Table \ref{tab:native-wer-value} illustrates that the ASR model's WER on audio from speakers whose first language is not English is twice as high as the WER on audio from native English speakers. To develop technologies that are inclusive to different speaker demographics, it is important to create benchmarks that are representative of these diverse backgrounds. 

%% file: 02_methods.tex
\begin{table*}[ht]
\caption{Comparing data statistics and different models compared on original and proposed splits for the Fluent Speech Commands dataset. Speaker and Utterance Coverage refer to the percentages of test set speakers and utterances, respectively, observed in the training set. ``Speaker KL" is the symmetrised Kullback-Leibler divergence of speaker demographic distributions between training and test sets. We ensure that our proposed splits have roughly the same \# of examples in each test set as in the split proposed in \cite{lugosch2019speech}. We also construct different variants of the Unseen split by changing the random seed of our algorithm and report the standard deviation. } 
\label{tab:model-comparison-fsc}
\resizebox {\linewidth} {!} {
  \centering
\begin{tabular}{llccc|cccccc}
\toprule
& \multicolumn{4}{c|}{Dataset Statistics} & \multicolumn{4}{c}{E2E SLU Model \cite{lugosch2019speech} Test Accuracy} \\ 
\cmidrule(r){2-9}
Fluent Speech & Speaker & Utterance & Speaker & Test & No & w/ Pretrained ASR & Finetune Word + & Finetune \\ 
% \cmidrule(r){3-4}\cmidrule(r){5-6}\cmidrule(r){8-9}\cmidrule(r){10-11}
Command Test Set & Coverage & Coverage & KL & Size & Pretraining & (Frozen) & Intent Layers & All Layers \\ 
 \midrule
Original Split & \hphantom{00}0\% & 100\% & 0.88 & 3793 & 96.8\hphantom{($\pm$0.4)} & 98.5\hphantom{($\pm$0.4)} & 99.1\hphantom{($\pm$0.4)} & 97.2\hphantom{($\pm$0.4)} \\
Random Split & 100\% & 100\% & \textless{0.01} & 3793 & 94.6\hphantom{($\pm$0.4)} & 96.2\hphantom{($\pm$0.4)} & 97.2\hphantom{($\pm$0.4)} & 95.8\hphantom{($\pm$0.4)} \\ \midrule
Unseen Split (Spk.) & \hphantom{00}0\% & 100\% & 0.01 & 3366 & 92.0 ($\pm$0.4) & 92.9 ($\pm$0.2) & 94.2 ($\pm$0.3) & 93.9($\pm$0.4)\\ 
Unseen Split (Utt.) & 100\% & \hphantom{00}0\% & \textless{0.01} & 3971 & 78.1 ($\pm$1.3) & 86.0 ($\pm$0.7) & 88.2 ($\pm$0.9) & 88.3($\pm$2.0)\\ 
\midrule
Challenge Split (Spk.) & \hphantom{00}0\% & 100\% & 0.01 & 3349 & 87.2\hphantom{($\pm$0.4)} & 90.9\hphantom{($\pm$0.4)} & 92.3\hphantom{($\pm$0.4)} & 91.1\hphantom{($\pm$0.4)}\\ 
Challenge Split (Utt.) & 100\% & \hphantom{00}0\% & \textless{0.01} & 4204 & 68.2\hphantom{($\pm$0.4)} & 73.4\hphantom{($\pm$0.4)} & 78.3\hphantom{($\pm$0.4)} & 74.1\hphantom{($\pm$0.4)}\\ 
\bottomrule
\end{tabular}
}
\vspace{-10px}
\end{table*}

% \begin{table}[ht]

% \caption{Comparing median and standard deviation values on the variants of the proposed Unseen split for the Fluent Speech Commands dataset} 
% \label{tab:model-comparison-fsc-variant}
% \resizebox {\linewidth} {!} {
%   \centering
% \begin{tabular}{lcccccccc}
% \toprule
% & \multicolumn{8}{c}{E2E SLU Model \cite{lugosch2019speech} Test Accuracy} \\ 
% \cmidrule(r){2-9}
% Fluent Speech & \multicolumn{2}{c}{No} & \multicolumn{2}{c}{w/ Pretrained ASR} & \multicolumn{2}{c}{Finetune Word +} & \multicolumn{2}{c}{Finetune} \\ 
% % \cmidrule(r){3-4}\cmidrule(r){5-6}\cmidrule(r){8-9}\cmidrule(r){10-11}
% Command Test Set & \multicolumn{2}{c}{Pretraining}& \multicolumn{2}{c}{(Frozen)} & \multicolumn{2}{c}{Intent Layers} & \multicolumn{2}{c}{All Layers} \\ 
%  \midrule
% Unseen Split (Spk.) & 91.5 & \textcolor{blue}{$\pm$0.4} & 93.3 & \textcolor{blue}{$\pm$0.2} & 94.3 & \textcolor{blue}{$\pm$0.3} & 93.7 & \textcolor{blue}{$\pm$0.4} \\ 
% Unseen Split (Utt.) & 77.4 & \textcolor{blue}{$\pm$1.3} & 86.0 & \textcolor{blue}{$\pm$0.7} & 88.2 & \textcolor{blue}{$\pm$0.9} & 86.5 & \textcolor{blue}{$\pm$2.0} \\ 
% \bottomrule
% \end{tabular}}
% \end{table}

%  & Unseen & \multicolumn{2}{c}{Challenge Split}
%  & Split & Split & Speaker & Utterance
%
% & 78.5 & 78.2 & 63.8 
% & XX & XX & XX
% & 75.0 & 86.4 & 71.1
% & 27.3 & 31.7 & 31.2

% \section{Quantifying Generalisability}
\section{Methodology} 
\label{CoordinateAscent}
% As a proxy for generalisability, we select task-specific \textit{utility functions} for the utterance-closed and speaker-closed scenarios.
% \begin{enumerate}
%     \item \textit{KL divergence of intent distribution} We minimize the KL-divergence between the train and test splits' intent distributions to ensure that both train and test set have similar semantic distribution. We also minimise the length distributions between train and test set.
%     \item \textit{N-gram overlap} BLEU score is used in NLP to compare word overlap between a source and target sentences. We minimize BLEU score (and thus word overlap) between utterances in the training/validation and test splits, providing a more realistic evaluation of a model's NLU capabilities.
%     \item \textit{KL-Divergence of Speaker Demographics.} FSC dataset comprises of speakers of various ages, English fluency levels, and different genders. We compute the distributions of each of these speaker demographic features. When generating closed-speaker training and test splits, we minimize the KL divergence between the distributions of speaker demographics, thus assuring we are testing for a variety of speaker demographics (compare to the \textit{Original} split, which contains all native English speakers).
% \end{enumerate}

% %\\\noindent\textbf{New Test Splits}\\
% \\\noindent 
%A robust SLU model should generalize to new speakers (thus, have powerful ASR capabilities) and to new utterances (have powerful NLU capabilities).
We discuss our approach for creating the open-utterance and open-speaker test sets for the \textit{Unseen} and Challenge splits. The \textit{Challenge} split uses additional constraints to make both test sets more difficult and realistic. 

\subsection{Dataset Optimization}
We construct test splits using coordinate ascent \cite{Wright2015CoordinateDA,Coordinate_Ascent} over sub-task driven utility functions. Each coordinate direction corresponds to the test set assignment (either the open-speaker or open-utterance test sets) of a block of datapoints in the dataset. We first select an open-speaker test set, then choose the open-utterance test set. Finally, we randomly distribute the remaining instances into training and validation sets, preserving the original size ratio and intent distributions of these sets \cite{lugosch2019speech}. %Moreover, we ensure the distribution of intent labels is roughly the same between train and test splits.

\subsection{Unseen Split}
\label{subsection:unseen-split}
We use the two functions to generate unseen-speaker and unseen-utterance splits with desirable qualities.

\paragraphHdTop{Unseen Speaker Set} The FSC dataset contains speakers of various ages, native languages, English fluency levels, and genders, but the original, open-speaker test set is not representative of these groups. To ensure we are testing on speakers of diverse backgrounds, we minimize the symmetrised Kullback-Leibler (KL) divergence  \cite{Symmetry_KL} between the discrete distributions of speaker demographics in the training and test sets.
%This is important because the FSC dataset contains speakers of various ages, native languages, English fluency levels, and genders, but the original test set does not include speakers from all these groups. We define demographics distributions over the aforementioned attributes.
%(compare to the \textit{Original} split, which contains all native English speakers).

\paragraphHdTop{Unseen Utterance Set}  When selecting unique utterances to hold out from training, we minimize the symmetrised KL divergence of the discrete intent label distributions between training and test sets. Utterances with the same intent are semantically similar, ensuring the semantic distributions of training and test sets match. We also minimize the KL divergence of the discrete distributions of transcript lengths between training and test sets. 

\subsection{Challenge Split}
\label{subsection:challenge-split}
In addition to the constraints defined in the previous section, we define speaker-specific and transcript-specific utility functions to quantify the ``hardness'' for each subtask. When optimized, these functions create more challenging, realistic held-out sets. Notably, these test sets may capture dataset outliers due to noisy recordings, labeling errors, or poorly aligned data. Thus, we recommend using them in addition to the Unseen splits.
%When optimized, these functions create more challenging, realistic held-out sets; these sets can evaluate model performance on  
The proposed utility functions are specific to spoken language tasks, but could be replaced with arbitrary task-specific objectives.

\paragraphHdTop{Challenge Speaker Set}
We compute the Word Error Rate, measured by insertion, deletion, and substitution errors, of Google's ASR model \cite{GoogleASR} to identify particularly challenging utterances. However, high WER is not always indicative of a reasonably hard example. According to previous work \cite{use_subs_Florian}, substitution errors reflect confusions in ASR systems. Alignment errors, indicated by an increase in insertion and deletion errors and a large deviation between these quantities, are a sign of poor data quality.  
%, including a high deviation between the number of insertion and deletion errors 
To produce a challenging speaker set without compromising data quality, we
% , we aim to maximize substitution errors in the test set while minimizing insertion and deletion errors 
use the following utility function, $U_{\text{WER}}$:
\begin{equation*}
    U_{\text{WER}}= S-\alpha~|I-D| - \beta~I - \gamma~D
\end{equation*}
where $S$, $I$ and $D$ refer to substitution, insertion, and deletion rates, respectively, such that we maximize $S$ while minimizing $I$, $D$ and their deviation, $|I-D|$. $\alpha$, $\beta$ and $\gamma$ are hyperparameters. We empirically observe $\alpha=0.05$, $\beta=0.05$ and $\gamma=0.4$ work well for the FSC dataset, producing a challenging split without compromising on test-set data quality.\\
\paragraphHdTop{Challenge Utterance Set}
A dataset with many unique n-grams makes the SLU task more difficult \cite{semcomplex2020}. Thus, we create splits that minimize the n-gram overlap between our train and test set.  We choose the Sentence BLEU \cite{papineni-etal-2002-bleu} score as a proxy for n-gram overlap and use it in the following utility function:
\begin{equation*}
U_{\text{BLEU}} = - BP*\exp(\sum_{i=1}^{4} \alpha_i \log(p_i))
\end{equation*}
where $p_i$ is the modified precision \cite{papineni-etal-2002-bleu} for each n-gram, $\alpha_i$ weighs the respective importance of the $i^\text{th}$-gram overlap, and \textit{BP} is the brevity penalty \cite{papineni-etal-2002-bleu} penalizing shorter sentences. Transcripts in the FSC dataset are 3-5 words in length, thus, considering only 1-gram and 2-gram overlap (i.e. $\alpha_1=0.5, \alpha_2=0.5, \alpha_3=0.0, \alpha_4=0.0$) worked well for holding out unique n-grams not seen during training.

Finally, to ensure both \textit{Unseen} and \textit{Challenge}  speaker sets test generalization only to new speakers (and not utterances), we maximize the n-gram overlap with each split's respective training utterances. As Table \ref{tab:model-comparison-fsc} shows, our constructed speaker test sets have 100\% n-gram overlap with their respective training sets. Unlike the original splits, in which test speakers are less demographically diverse than training set speakers, our new splits effectively minimize this distributional gap, as shown by the ``Speaker KL" column. Each new test set has similar size to the original test set, and its distribution of utterance lengths is kept close to that of the training set to limit distribution shift.

%% file: 03_results.tex
\definecolor{blueone}{RGB}{237,248,251}
\definecolor{bluetwo}{RGB}{178,226,226}
\definecolor{bluethree}{RGB}{102,194,164}
\definecolor{bluefour}{RGB}{35,139,69}

\section{Experiments}

% \begin{table*}[h]
%   \centering
% \resizebox {\linewidth} {!} {
% \begin{tabular}{lccccccccc}
% \toprule
% &\multicolumn{4}{c}{Speaker Closed} & &  \multicolumn{4}{c}{Utterance Closed}  \\ 
% \cmidrule{2-5} \cmidrule{7-10}
%  & \multicolumn{2}{c}{\# Unique} & \multicolumn{2}{c}{SLU Model} & & \multicolumn{2}{c}{\# Unique} & \multicolumn{2}{c}{SLU Model} \\ 
% Split & \#Speakers & \#Utt. & Pretrained ASR & Finetune ASR & & \#Speakers & \#Utt. & Pretrained ASR & Finetune ASR \\ \midrule
% Original & 10 & 0 & 0.98 & 0.97 & & - & - & - & -  \\
% Closed Set & 13 & 0 & 0.93 & 0.93 &  & 0 & 40 & 0.86 & 0.88  \\
% +WER/BLEU & 13 & 0 & 0.93 & 0.93 &  & 0 & 39 & 0.75 & 0.75  \\
% \bottomrule
% \end{tabular}
% }
% \caption{Test accuracy of the state of the art SLU models on the original and proposed splits. We also mention the number of speakers and utterances that were unique in test set i.e. not seen in training set for given split}
% \label{table:slu-results}
% \end{table*}

\subsection{Comparing end-to-end SLU systems}
\label{subsection:comparing-e2e}
We compare four different models from \cite{lugosch2019speech} on the \textit{Original}, \textit{Unseen}, and \textit{Challenge} splits, as well as a stratified \textit{Random} split (stratified over all intent labels). The four models are based on a three-stage neural architecture consisting of a phoneme layer, word layer, and intent layer. Each model uses different pretraining and finetuning schemes: using no pretraining, using a frozen pretrained ASR model (i.e. finetuning only the intent layers), finetuning only word and intent layers, or finetuning all layers. When pretraining, the phoneme and word modules are pretrained on the LibriSpeech dataset \cite{Librispeech}. Using the Original test split, we successfully reproduced the results \cite{lugosch2019speech} for each of these freezing and unfreezing schedules. 
%Using the frozen pretrained ASR model or finetuning both word and intent layers yields the best performance.
%, In our attempt to reproduce their reported results, we observe our model performances to be very similar to those reported in \cite{lugosch2019speech} with finetuning only the intent layer and finetuning both word and intent layers giving best performance on the original test set.

Using the speaker and utterance test sets we create, we can highlight sub-task-level performance differences across the four models. As Table \ref{tab:model-comparison-fsc} illustrates, our Unseen and Challenge splits reveal that all models are better at generalizing to new speakers than to new utterances. However, all models achieve at least 3\% lower accuracy on the Unseen and Challenge speaker sets compared with the Original held-out speaker set, indicating that current SLU models still do not generalize well to diverse speaker demographics.
%These performance gaps are not seen with the \textit{Original} test split, on which all models achieve very similar performance with accuracies greater than 95\%.
The results on the Challenge utterance set indicate that all models are significantly worse at generalizing to unique phrases of the same intent, suggesting an opportunity for enhancing semantic processing abilities of SLU models.
%With the \textit{Unseen} and \textit{Challenge} sets, we observe consistently larger performance gaps between these models. We also observe worse performance from all models when evaluated on the utility-optimized challenge sets.

Our splits are also useful for comparing configurations of the same model. Intuitively, pretraining phoneme layers to detect phonetic patterns should help generalize to unseen speakers and utterances.  However, Table \ref{tab:model-comparison-fsc} shows that on the Original and Random splits, there are small performance gaps between pretrained and non-pretrained models (1-2\%, or $\sim$50 test set examples), suggesting pretraining offers limited value, considering the resources it requires. In contrast, the performance gap becomes significant in the Unseen utterance set and both the speaker and utterance Challenge splits. The model without pretraining performs 10.1\% worse than the best pre-trained model (pretraining with finetuned word and intent layers) in both the Unseen and Challenge utterance sets, corresponding to $\approx$460 more mistakes. The gaps are smaller in the speaker Challenge set, suggesting a non-pretrained ASR model generalizes better to new speakers than to new words or phonemes. 
%In contrast to the results reported in \cite{lugosch2019speech}, we interestingly find that finetuning more layers gives better performance on our proposed test sets, with finetuning word and intent layers being consistently the best. 
These results corroborate previous findings \cite{semcomplex2020} that finetuning models to the dataset's distinct acoustic and linguistic patterns improves generalization to new phrasings. 
Finally, we change the random seeds used to create the Unseen split to test the robustness of our methods. The relatively low standard deviations in performance, as seen in Table \ref{tab:model-comparison-fsc}, illustrate that our method is stable.
%Moreover, we find that the non-pretrained model is the worst on all evaluations, and observe particularly large deficiencies when evaluating on unseen utterances. 

% \begin{table}[h]
% \resizebox {\linewidth} {!} {
% \begin{tabular}{llc}
% \toprule
% Embeddings & \multicolumn{1}{c}{Test Split} & \multicolumn{1}{l}{Test Accuracy} \\ \midrule
% \multirow{3}{*}{Rand. Init. (With Finetune)} & Original Set & 98.1  \\   
%  & Unseen Split & 90.3 \\
%  & Challenge Split & 78.8 \\ \midrule
% \multirow{3}{*}{with FastText} & Original Set & 98.1  \\  
%  & Unseen Split & 89.8 \\
%  & Challenge Split & 83.7 \\ \hline
% \multirow{3}{*}{with BERT (Pretrained)} & Original Set & 98.1 \\ 
%  & Unseen Split & 87.9 \\ 
%  & Challenge Split & 86.1 \\ \hline
% \multirow{3}{*}{with BERT (Finetuned)} & Original Set & 98.1 \\ 
%  & Unseen Split & 90.4 \\ 
%  & Challenge Split & 86.1 \\ \bottomrule
% \end{tabular}}
% \caption{Test accuracies of NLU-only models using different embedding types, presented on the original and proposed splits (on the unseen utterance test set)\sd{change to \% values and 3 decimal}}
% \label{tab:nlu-only}
% \end{table}

% xlabel={Test Splits},
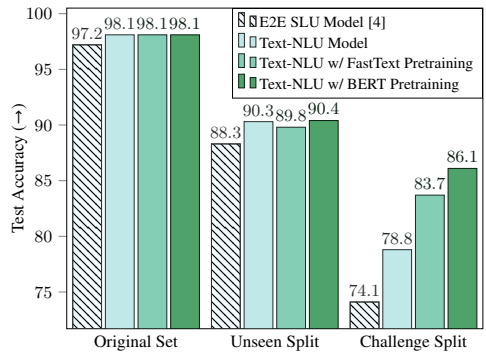
\begin{figure}[t]
\centering
\resizebox{0.8\linewidth}{!}{
    \begin{tikzpicture}
        \begin{axis}[
            ybar,
            enlarge x limits=0.25,
            bar width=15pt, 
            ylabel={Test Accuracy ($\rightarrow$)},
            symbolic x coords={Original Set, Unseen Split, Challenge Split},
            xtick=data,
            nodes near coords,
            nodes near coords align={vertical},
            legend cell align={left},
            xlabel near ticks,
            ylabel near ticks,
            ytick align=outside,
            ytick pos=left,
            xtick align=inside,
            xtick style={draw=none},
            ylabel shift=-7 pt,
            height=7.5cm,
            width=9.2cm,
            legend style={at={(1,1)},anchor=north east,nodes={scale=0.9, transform shape}}
            ]
    %         \addlegendimage{empty legend}\addlegendentry{\hspace{-.4cm}\underline{SLU System}}
    % 		\addlegendentry{SLU}
    % 		\addlegendimage{empty legend}\addlegendentry{\hspace{-.4cm}\underline{NLU System}}
    % 		\addlegendentry{Random}
    % 		\addlegendentry{FastText}
    % 		\addlegendentry{BERT}
            \addplot[blueone!20!black, fill=blueone!80!white!, , postaction={
        pattern=north west lines
    }] coordinates {(Original Set,97.2) (Unseen Split,88.3) (Challenge Split,74.1)};
            \addplot[bluetwo!20!black,fill=bluetwo!80!white] coordinates {(Original Set,98.1) (Unseen Split,90.3) (Challenge Split,78.8)};
            \addplot[bluethree!20!black,fill=bluethree!80!white] coordinates {(Original Set,98.1) (Unseen Split,89.8) (Challenge Split,83.7)};
            \addplot[bluefour!20!black,fill=bluefour!80!white] coordinates {(Original Set,98.1) (Unseen Split,90.4) (Challenge Split,86.1)};
            \legend{E2E SLU Model \cite{lugosch2019speech}, Text-NLU Model, Text-NLU w/ FastText Pretraining, Text-NLU w/ BERT Pretraining}
        \end{axis}
        
    \end{tikzpicture}
    }
    \caption{Comparing text-based NLU models with the ``Finetune All Layers SLU" baseline on the original and proposed utterance test sets. ``Text-NLU Model" refers to a text-based NLU system using randomly initialized word embeddings.}
    \label{fig:nlu-only}
    \vspace{-10px}
\end{figure}
            % \legend{E2E SLU Model \cite{lugosch2019speech}, Text-NLU (Rand. Init), Text-NLU (FastText Pretrained), Text-NLU (BERT Pretrained)}

\subsection{Gap between SLU and NLU}
Using the utterance test sets of the Unseen and Challenge splits, we identified that SLU systems struggle to effectively capture lexical and semantic information. As an ablation study, we used gold transcripts to train and test the intent prediction component of the end-to-end model \cite{lugosch2019speech} in isolation (keeping all word and phoneme layers frozen). As a baseline, we train a text-based intent classification model initialized with random word embeddings that are finetuned during training.   
%and capture semantic information by extracting word vectors. Using our utterance-closed splits, we can  identified the gap of SLU systems in capturing lexical and semantic information,
To incorporate semantic information into the word representations, we extract two types of word embeddings: (1) pretrained FastText \cite{mikolov2018advances} embeddings and (2) contextual BERT embeddings \cite{devlin2018bert}. In Figure \ref{fig:nlu-only}, we compare the baseline and semantically-enhanced text NLU models with the ``Finetune All Layers" SLU model of \cite{lugosch2019speech}.
%We use these three embedding types as input for training the intent sub-component; for the FastText and BERT settings, we additionally experiment with finetuning the embeddings to our task.
%Compared with NLU models, the SLU model underperforms on both the Challenge and Unseen splits, implying that the intent module also has a significant gap to bridge.
Figure \ref{fig:nlu-only} illustrates that BERT pretraining can boost the accuracy of the intent subcomponent by 12\% on the Challenge utterance set. These differences are not so apparent in the Unseen split, which is not as semantically challenging because it does not explicitly minimize n-gram overlap \cite{semcomplex2020}. There is still a 2\% gap between the SLU and NLU models' performance on the Unseen utterance split, suggesting that pretraining embeddings helps enhance semantic understanding.

\begin{table}[t]
\caption{Adding semantic word embeddings to the SLU system has only a minor effect ($<$1\%) on the the Original split and proposed unseen-speaker splits. On the unseen-utterance splits, we see a magnified performance gap ($>$2\%), in \textbf{bold}.} 
\label{tab:model-comparison-fasttext}
 \centering
 \resizebox {\linewidth} {!} {
\begin{tabular}{lc|cc|cc}
\toprule
&  & \multicolumn{2}{c|}{Unseen} & \multicolumn{2}{c}{Challenge} \\
E2E SLU Model \cite{lugosch2019speech} & Original & Spk. & Utt. & Spk. & Utt.  \\ 
% \cmidrule(r){3-4}\cmidrule(r){5-6}\cmidrule(r){8-9}\cmidrule(r){10-11}
 \midrule
Pretrained ASR (Frozen) & 98.5 & 92.9 & 86.0 & 90.9 & 73.4\\
% \midrule
+ FastText Pretraining & 98.7 & 92.7 & \textbf{88.3} & 90.0 & \textbf{75.5}   \\ 
\bottomrule
\end{tabular}
}
% \vspace{-10px}
\end{table}

Based on the results of our ablation study, we extend the frozen pretrained ASR model, the best reported SLU model from \cite{lugosch2019speech}, with FastText embeddings. At each audio frame, the ASR module predicts a distribution over words; we use this compute a weighted average FastText word embedding \cite{bojanowski-etal-2017-enriching} and pass it to the intent layer. Table \ref{tab:model-comparison-fasttext} illustrates that enriching the ASR outputs with semantic information gives very minor improvements on the original test set, but provides $>$2\% improvement to the unseen-utterance sets of both Unseen and Challenge splits, consistent with the experiments in the previous section. 
% We also observe slightly negative effects on our diverse speaker test sets; additional semantic information may hurt performance if predicted top-k words are incorrect.

\subsection{Analyzing proposed utility functions}
In Section \ref{subsection:comparing-e2e}, we illustrated how our optimized splits can distinguish model performance. We now verify our utility functions can effectively quantify the complexity of each subtask.
% \small{
% \begin{table}[]
% \footnotesize
% \begin{tabular}{cllll}
% \hline
% \multirow{2}{*}{Test Split} & Insertion & Substitution & Deletion & WER \\
%  & \multicolumn{1}{l}{\begin{tabular}[c]{@{}l@{}}U/S\end{tabular}} & \multicolumn{1}{l}{\begin{tabular}[c]{@{}l@{}}U/S\end{tabular}}
%  & \multicolumn{1}{l}{\begin{tabular}[c]{@{}l@{}}U/S\end{tabular}}
%  & \multicolumn{1}{l}{\begin{tabular}[c]{@{}l@{}}U/S\end{tabular}}
%  \\ \hline
% Original &  0.005 & 0.010 & 0.006 & 0.065 \\ \hline
% Closed Set & 0.010/0.006 & 0.021/0.019 & 0.025/0.023 & 0.121/0.093 \\ \hline
% \multirow{1}{*}{\begin{tabular}[c]{@{}c@{}}Speaker-Utility, \\ Min N-gram Overlap\end{tabular}} & 0.010/0.006 & 0.020/0.024 & 0.024/0.024 & 0.118/0.112 \\
% & & \\ \hline
% \end{tabular}
% \caption{WER value for test set in the original and proposed splits}
% \label{tab:wer-value}
% \end{table}}

\begin{table}[t]
\caption{\% WER values of the Google ASR system \cite{GoogleASR} on the original and proposed speaker test sets, and the corresponding accuracy of the Pretrained ASR (Frozen) model. (S, I, and D refers to substitution, insertion, and deletion,  respectively.)}
\label{tab:wer-value}
  \centering
\resizebox {\linewidth} {!} {
\begin{tabular}{lccccc}
\toprule
Test Split     & \% S   & \% I   & \% D   & \% WER & SLU Acc.\\ \midrule
Original       & 1.0 & 0.5 & 0.6 & \hphantom{0}6.5 & 98.5 \\ 
Unseen Speaker & 2.1 & 1.0 & 2.5 & 12.1  & 92.9\\ 
Challenge Speaker  & 3.2 & 1.2 & 2.6 & 13.9 & 90.9 \\
\bottomrule
\end{tabular}
}
\vspace{-10px}
\end{table}

\paragraphHdTop{Word Error Rate} As in Section \ref{subsection:task}, we use WER of Google's ASR system \cite{GoogleASR} to quantify the difficulty of our splits. Table \ref{tab:wer-value} illustrates that WER is twice as high on the proposed speaker-diverse splits as on the original splits. Substitution errors are most prominent in Challenge set, indicating that we create a hard test set without necessarily compromising on data quality.\\
\paragraphHdTop{N-gram overlap} For each proposed split, we compute the average BLEU score for the test set relative to the training set. Table \ref{tab:ngram-value} highlights that our test splits have much lower n-gram overlap with their training sets. Minimizing n-gram overlap while preserving intent distributions of training and test sets further tests a model's generalization to new phrasings of the same intents.  

% \small{
% \begin{table}[]
% \footnotesize
% \begin{tabular}{cllll}
% \hline
% \multirow{2}{*}{Test Split} & BLEU & BLEU & BLEU & BLEU \\
%  & \multicolumn{1}{l}{\begin{tabular}[c]{@{}l@{}}(1)\end{tabular}} & \multicolumn{1}{l}{\begin{tabular}[c]{@{}l@{}}(0,1)\end{tabular}}
%  & \multicolumn{1}{l}{\begin{tabular}[c]{@{}l@{}}(0,0,1)\end{tabular}}
%  & \multicolumn{1}{l}{\begin{tabular}[c]{@{}l@{}}(0,0,0,1)\end{tabular}}
%  \\ \hline
% Original &  1.0 & 1.0 & 1.0 & 1.0 \\ \hline
% Closed Set & 1.0/0.996 & 1.0/0.966 & 1.0/0.904 & 1.0/0.717 \\ \hline
% \multirow{1}{*}{\begin{tabular}[c]{@{}c@{}}Speaker-Utility, \\ Min N-gram Overlap\end{tabular}} & 1.0/0.901 & 1.0/0.664 & 1.0/0.680 & 1.0/0.712 \\
% & & \\ \hline
% \end{tabular}
% \caption{BLEU score values for only unigram (1), bigram (0,1), trigram (0,0,1) and 4-gram (0,0,0,1). To compute these statistics, those utterances with length less than order of n-gram in consideration were removed.  }
% \label{tab:wer-value}
% \end{table}}
\begin{table}[t]
\caption{BLEU score values for unigram, bigram, trigram and 4-gram for Original and proposed (utterance) test sets. Utterances shorter than order of a given n-gram were removed.}
\label{tab:ngram-value}
  \centering
  \resizebox {\linewidth} {!} {
\footnotesize
\begin{tabular}{lccccc}
\toprule
& \multicolumn{4}{c}{N-gram overlap} & SLU \\ 
\cmidrule{2-5} 
Test Split & 1 & 2 & 3 & 4 & Acc.\\
\midrule
Original &  100.0 & 100.0 & 100.0 & 100.0 & 98.5 \\ 
Unseen Utterance & \hphantom{0}98.0 & \hphantom{0}87.4 & \hphantom{0}73.4 & \hphantom{0}71.7 & 86.0 \\ 
Challenge Utterance & \hphantom{0}91.0 & \hphantom{0}69.9 & \hphantom{0}66.4 & \hphantom{0}66.6 & 73.4\\
\bottomrule
\end{tabular}}
% \vspace{-10px}
\end{table}

% \subsection{Examining performance differences between models \sd{part of the 4.1} }
% One goal of our test splits is to magnify performance differences between models. To this end, we compare performance of 5 different models on both the original FSC splits and on our newly generated splits. 

\subsection{Extending to the Snips SmartLights dataset}

To illustrate our methodology is dataset agnostic, we extend our approach to Snips SmartLights, a popular SLU dataset \cite{Lugosch2020UsingSS,Bhosale2019EndtoEndSL}. 

\begin{table}[t]
\caption{Evaluating models on original and proposed splits for the Snips SmartLights dataset. Snips does not provide default splits, so we compare against a random split.} 
\label{tab:model-comparison-snips}
\resizebox {\linewidth} {!} {
  \centering
\begin{tabular}{lc|c|cc}
\toprule
% & \multicolumn{4}{c}{Snips SmartLights Dataset} \\ 
% \cmidrule(r){2-5}
 & Rand. \cite{Agrawal2020TieYE} & Unseen & \multicolumn{2}{c}{Challenge Split}\\ 
% \cmidrule(r){3-4}\cmidrule(r){5-6}\cmidrule(r){8-9}\cmidrule(r){10-11}
E2E SLU Model \cite{lugosch2019speech} & Split & Split & (Spk.) & (Utt.) \\ \midrule
No Pretraining & 60.4 & 27.3 & 37.8 & 45.2  \\
w/ Pretrained ASR (Frozen) & 83.2 & 78.5 & 73.2 & 67.4  \\ 
Finetune Word + Intent Layers & 88.0 & 80.9 & 82.6 & 75.3 \\ 
Finetune All Layers & 85.0 & 75.0 & 74.8 & 78.5  \\ 
\bottomrule
\end{tabular}
}
\vspace{-10px}
\end{table}

Snips SmartLights dataset is unseen-utterance by design because all utterances are unique. Thus, we create a single Unseen test set that holds out speakers and utterances. We optimize both speaker and utterance utilities defined in Section \ref{subsection:unseen-split} to create the split.  Using the WER and n-gram based utilities defined in Section \ref{subsection:challenge-split}, we create separate speaker and utterance Challenge test sets.
Following the Challenge test setup of FSC, we increase the speaker test set's n-gram overlap set with its train set to match that of the random split. We do not control for n-gram overlap in the Unseen split since it holds out both speakers and utterances. As a result, our Challenge speaker set may be easier than the Unseen split for SLU models. Moreover, Snips is a smaller dataset, so the Snips Challenge set's train, valid, speaker test, utterance test ratios are 75:10:7.5:7.5 as compared to 80:10:10 in the baseline random split \cite{Agrawal2020TieYE}.

Using the same models as Section \ref{subsection:comparing-e2e}, we compare the performance on our proposed splits against a random split \cite{Agrawal2020TieYE} in Table \ref{tab:model-comparison-snips} (Snips does not release official splits). We observe similar results as in the FSC setting. 
%The Pretrained ASR (Frozen) model observes a significant drop in performance on our optimized splits. 
Pretraining ASR models improves speaker test set performance by nearly 40-50\% for the Unseen and Challenge splits. Moreover, finetuning improves performance for all splits, especially on the Challenge utterance set, on which both finetuned models achieve nearly 12\% gains in performance. Thus, we illustrate that we can easily extend our approach to another SLU benchmark, and see effects consistent with those on the FSC dataset. 

%% file: 04_conclusion.tex
\section{Conclusions}
We present a novel, dataset-agnostic methodology for constructing splits for decomposable tasks, casting the construction of splits as an optimization problem over dataset-level utility functions. We release \emph{Unseen} and \emph{Challenge} splits for the FSC and Snips datasets to the community, and show evidence that these splits can amplify performance differences between sub-components of models. We recommend the use of the \emph{Unseen} splits for testing in-domain performance and the \emph{Challenge} splits for more extreme out-of-domain generalization scenarios. As our methodology is task-agnostic, we encourage the extension of our re-splitting method to other decomposable tasks, such as speech translation or visual question answering. 
\section{Acknowledgements}
This work was supported in part by the National Science Foundation under Grant No. IIS2040926, the NSF SaTC Frontier project(CNS-1914486) \cite{Norman1}, Bridges PSC (ACI-1548562, ACI-1445606) and an AWS Machine Learning Research Award. 